# What Cannot be Learned with Bethe Approximations

Uri Heinemann    Amir Globerson
School of Computer Science and Engineering
The Hebrew University, Jerusalem 91904, Israel

## Abstract

We address the problem of learning the parameters in graphical models when inference is intractable. A common strategy in this case is to replace the partition function with its Bethe approximation. We show that there exists a regime of empirical marginals where such Bethe learning will fail. By failure we mean that the empirical marginals cannot be recovered from the approximated maximum likelihood parameters (i.e., moment matching is not achieved). We provide several conditions on empirical marginals that yield outer and inner bounds on the set of Bethe learnable marginals. An interesting implication of our results is that there exists a large class of marginals that cannot be obtained as stable fixed points of belief propagation. Taken together our results provide a novel approach to analyzing learning with Bethe approximations and highlight when it can be expected to work or fail.

Probabilistic graphical models [8, 23] are a powerful tool for describing complex multivariate distributions. They have been used successfully in a wide range of fields, from computational biology to machine vision and natural language processing. To use such a model in practice, one typically needs to solve two related tasks. The first is the inference task which involves calculating probabilities of events under the model. The second task involves learning the parameters of the model from empirical data.

Unfortunately, in many models of interest the inference problem is computationally hard, and cannot be solved exactly in practice. This has motivated extensive research into approximate inference schemes, some of which have been quite successful empirically. Perhaps the most well known of these is the belief propagation (BP) algorithm, which is closely related to variational approximations based on Bethe free energies [26]. Another variational approach, which uses convex free energies is the tree-reweighted (TRW) method [22]. Although the TRW approach results in convex optimization problems for inference, it sometimes yields marginals that are inferior to those obtained by BP (e.g., see [9]).

How should one learn the parameters of a model when inference is intractable? The typical approach to parameter learning is likelihood maximization, but when inference is intractable it is also hard to maximize the likelihood.[1] Because of this difficulty, many methods have been devised to approximate the learning problem. One elegant approach is to approximate the likelihood using the same variational approximation that is employed during inference [5, 14, 16, 19].

Analyzing the performance of approximate learning schemes is challenging, since even the accuracy of the underlying variational approximations is hard to analyze. Furthermore, we do not generally expect the learned model to be similar to the one obtained using exact maximum likelihood. One approach, which has recently been introduced by Wainwright [19] is to use the notion of moment matching. In exact maximum likelihood learning, the learned model has a nice property: some if its marginals are guaranteed to be identical to those of the empirical data. This property is often referred to as *moment matching*. Wainwright [19, 21] has shown that when using convex variational approximations such as TRW, the learned model also has the moment matching property in the following sense: if one applies approximate inference to it (using the same variational approach that was used during learning), the resulting marginals will be equal to the empirical ones. However, these results cannot be applied to learning with Bethe approximations, since the latter are not convex. Because of the success of Bethe approximations in a wide array of applications, it is important to understand the advantages and limitations of learning with those. This is precisely the goal of our work.

---

[1] When the data are known to be generated by a graphical model of the same structure, pseudo-likelihood [1] can be used and is consistent. However, this assumption is rarely met in practice, and pseudo-likelihood often does not perform well in these cases.

It may initially seem like learning with Bethe approximations would also result in a moment matching property. In other words, if we use Bethe approximations during both learning and inference, our learned model will agree with the empirical marginals. However, as we show here, the situation is considerably more complex. In the current work we provide some surprising results with respect to moment matching and Bethe approximations, that shed light on the performance of learning with such approximations, and on properties of the BP algorithm. Our main results are:

- We show that there exist empirical distributions for which Bethe approximations *cannot* perform moment matching. In other words, if we run BP on the optimal Bethe parameters, we will not recover the empirical marginals. Such empirical distributions are thus *bad inputs* for Bethe approximations, since the learned parameters cannot be used to reconstruct the original marginals.

- We provide inner and outer bounds on the set of marginals for which Bethe moment matching is possible, and show that they agree with empirical behavior of Bethe learning. Surprisingly, we show that binary attractive models cannot be learned with Bethe approximations for certain graphs.

- Our results also provide a novel characterization of BP fixed points. Specifically, we show that there is a large class of marginals that cannot be obtained as stable fixed points of BP.

Taken together, our results provide a novel way of analyzing learning with Bethe approximations.

## 1 Maximum Likelihood in Graphical Models

We focus on pairwise Markov random fields for simplicity. That is, we consider random variables $X_1, \ldots, X_{N_V}$ and pairwise functions $\theta_{ij}(x_i, x_j)$ corresponding to edges $E$ in a graph $G$ with $N_V$ nodes. The MRF corresponding to these parameters is given by:

$$p(\boldsymbol{x}; \boldsymbol{\theta}) = \frac{1}{Z(\boldsymbol{\theta})} \exp\left(\sum_{ij \in E} \theta_{ij}(x_i, x_j) + \sum_{i=1}^{N_V} \theta_i(x_i)\right) \quad (1)$$

where $Z(\boldsymbol{\theta})$ is the partition function and $\boldsymbol{x}$ corresponds to a complete assignment to the $N_V$ variables.

We wish to learn the parameters $\boldsymbol{\theta}$ from a sample of size $M$ given by $\boldsymbol{x}^{(1)}, \ldots, \boldsymbol{x}^{(M)}$. A standard approach to parameter learning is to maximize the log likelihood given by:[2]

$$\ell(\boldsymbol{\theta}) = \frac{1}{M} \sum_m \log p(\boldsymbol{x}^{(m)}; \boldsymbol{\theta}) = \bar{\boldsymbol{\mu}} \cdot \boldsymbol{\theta} - \log Z(\boldsymbol{\theta}) \quad (2)$$

where $\bar{\boldsymbol{\mu}}$ are the empirical marginals given by:

$$\bar{\boldsymbol{\mu}}_{ij}(x_i, x_j) = \frac{1}{M} \sum_m \delta_{x_i^m, x_i} \delta_{x_j^m, x_j} \quad \forall ij \in E$$

$$\bar{\boldsymbol{\mu}}_i(x_i) = \frac{1}{M} \sum_m \delta_{x_i^m, x_i} \quad i = 1, \ldots, N$$

and $\boldsymbol{\theta}$ is the corresponding vector with the parameters $\theta_{ij}(x_i, x_j), \theta_i(x_i)$ in appropriate coordinates.

The likelihood function $\ell(\boldsymbol{\theta})$ is concave in $\boldsymbol{\theta}$ and thus does not have local optima. Finding its global maximum is possible when the function, as well as its gradient, can be calculated efficiently. In these cases a variety of methods can be used, such as gradient ascent, iterative proportional fitting or any other general purpose first order convex optimization procedure.

A key property of the optimal parameters $\boldsymbol{\theta}_{ML}(\bar{\boldsymbol{\mu}})$ is that they satisfy the so called moment matching condition, described next. Define the vector $\boldsymbol{\mu}^{\boldsymbol{\theta}}$ to be the set of marginals of the model $p(\boldsymbol{x}; \boldsymbol{\theta})$ given by:

$$\boldsymbol{\mu}_{ij}^{\boldsymbol{\theta}}(x_i, x_j) = p(x_i, x_j; \boldsymbol{\theta}) \quad \forall ij \in E$$
$$\boldsymbol{\mu}_i^{\boldsymbol{\theta}}(x_i) = p(x_i; \boldsymbol{\theta}) \quad i = 1, \ldots, N$$

The moment matching condition for maximum likelihood optimality is then simply:

$$\boldsymbol{\mu}^{\boldsymbol{\theta}_{ML}} = \bar{\boldsymbol{\mu}} \quad (3)$$

The condition states the following simple fact: the optimal parameters $\boldsymbol{\theta}_{ML}$ are such that the optimal model $p(\boldsymbol{x}; \boldsymbol{\theta}_{ML})$ has the given empirical marginals. This is a desirable property since the empirical marginals are often a good approximation of the true marginals.[3]

### 1.1 Bethe Approximations of the Likelihood

The problem of maximizing the likelihood in Eq. 2 is hard due to the general intractability of the partition function and marginals inference problems. We shall focus on a common approach to this problem, which is to replace $\log Z(\boldsymbol{\theta})$ with an approximation $F(\boldsymbol{\theta})$. We shall specifically be interested in the Bethe approximation [26] defined as follows. First define the negative Bethe free energy as:

$$F(\boldsymbol{\mu}; \boldsymbol{\theta}) = \boldsymbol{\mu} \cdot \boldsymbol{\theta} + H_B(\boldsymbol{\mu}) \quad (4)$$

---

[2]Dependence on the sample is implicit throughout.

[3]When not enough data is available for estimating marginals reliably from data, it is possible to use smoothing or regularization. We do not address this here, but our results can be generalized to these cases.

Where the function $H_B(\boldsymbol{\mu})$ is the Bethe entropy given by:

$$H_B(\boldsymbol{\mu}) = \sum_{i \in V}(d_i - 1) \sum_{x_i} \mu_i(x_i) \log \mu_i(x_i)$$
$$- \sum_{ij \in E} \sum_{x_i, x_j} \mu_{ij}(x_i, x_j) \log \mu_{ij}(x_i, x_j)$$

and $d_i$ is the degree of node $i$ in the graph $G$.

The Bethe approximation for the log partition function $\log Z(\boldsymbol{\theta})$ is then given by:

$$F(\boldsymbol{\theta}) = \max_{\boldsymbol{\mu} \in \mathcal{M}_L} F(\boldsymbol{\mu}; \boldsymbol{\theta}) \quad (5)$$

Eq. 5 uses the following definitions: $\mathcal{M}_L$ is the *local marginal polytope*, which is the set of locally consistent pseudo-marginals $\boldsymbol{\mu}$ defined as:

$$\mathcal{M}_L = \left\{ \boldsymbol{\mu} \geq 0 \, \middle| \, \begin{array}{l} \sum_{x_j} \mu_{ij}(x_i, x_j) = \mu_i(x_i) \;\; ij \in E \\ \sum_{x_i} \mu_{ij}(x_i, x_j) = \mu_j(x_j) \;\; ij \in E \\ \sum_{x_i} \mu_i(x_i) = 1 \end{array} \right\}. \quad (6)$$

We are thus interested in maximizing the Bethe likelihood:

$$\ell_B(\boldsymbol{\theta}; \bar{\boldsymbol{\mu}}) = \bar{\boldsymbol{\mu}} \cdot \boldsymbol{\theta} - F(\boldsymbol{\theta}) = \bar{\boldsymbol{\mu}} \cdot \boldsymbol{\theta} - \max_{\boldsymbol{\mu} \in \mathcal{M}_L} F(\boldsymbol{\mu}; \boldsymbol{\theta}) \quad (7)$$

One interesting property of the function $\ell_B(\boldsymbol{\theta}; \bar{\boldsymbol{\mu}})$, which is typically overlooked, is that it is in fact a *concave* function of $\boldsymbol{\theta}$ and thus it does not have local maxima. This is a simple outcome of the fact that $F(\boldsymbol{\theta})$ is a pointwise maximum over functions that are convex (in fact linear) in $\boldsymbol{\theta}$ and is thus convex [2], so that its negation is concave.

In what follows, we characterize the global maximizer of the Bethe likelihood and discuss several ways of approximating it. We denote the maximizer by $\boldsymbol{\theta}_B(\bar{\boldsymbol{\mu}})$.

### 1.2 Optimality Conditions for Bethe Learning

The Bethe likelihood $\ell_B(\boldsymbol{\theta}; \bar{\boldsymbol{\mu}})$ is a concave but non-smooth function. Thus, a necessary and sufficient condition for $\boldsymbol{\theta}_B(\bar{\boldsymbol{\mu}})$ to maximize it is that the subdifferential of $\ell_B(\boldsymbol{\theta}; \bar{\boldsymbol{\mu}})$ at $\boldsymbol{\theta}_B(\bar{\boldsymbol{\mu}})$ contains the all zero vector $\boldsymbol{0}$ [12]. In what follows we use this to characterize $\boldsymbol{\theta}_B(\bar{\boldsymbol{\mu}})$.

The subdifferential of $F(\boldsymbol{\theta})$ is defined as follows. For a given $\boldsymbol{\theta}$ define $M(\boldsymbol{\theta})$ to be the set of vectors $\boldsymbol{\mu}$ that maximize $F(\boldsymbol{\mu}; \boldsymbol{\theta})$. Namely:

$$M(\boldsymbol{\theta}) = \arg \max_{\boldsymbol{\mu} \in \mathcal{M}_L} F(\boldsymbol{\mu}; \boldsymbol{\theta}) \quad (8)$$

The subdifferential of $F(\boldsymbol{\theta})$ is then $\mathrm{C}onv\{M(\boldsymbol{\theta})\}$, the convex hull of the vectors in $M(\boldsymbol{\theta})$. The subdifferential of $\ell_B(\boldsymbol{\theta}; \bar{\boldsymbol{\mu}})$ is thus $\bar{\boldsymbol{\mu}} - \mathrm{C}onv\{M(\boldsymbol{\theta})\}$. Taken together we have that the optimality condition for $\boldsymbol{\theta}_B(\bar{\boldsymbol{\mu}})$ is:

$$\bar{\boldsymbol{\mu}} \in \mathrm{C}onv\{M(\boldsymbol{\theta}_B(\bar{\boldsymbol{\mu}}))\} \quad (9)$$

Whenever there is a single maximizer $\boldsymbol{\mu}(\theta)$ in $M(\boldsymbol{\theta})$ the condition becomes $\bar{\boldsymbol{\mu}} = \boldsymbol{\mu}(\theta)$, resembling the moment matching condition in Eq. 3. However, generally the above condition is more subtle and actually means that when there are multiple maximizers, $\bar{\boldsymbol{\mu}}$ is not necessarily equal to any of them. In Section 2 we consider the strong implications that this has on learning with Bethe approximations.

### 1.3 Maximizing the Bethe Likelihood in Practice

Although $\ell_B(\boldsymbol{\theta}; \bar{\boldsymbol{\mu}})$ is concave, finding its maximum is still hard and there is no known closed form solution or polynomial time algorithm for it. The key difficulty is that both evaluating $\ell_B(\boldsymbol{\theta}; \bar{\boldsymbol{\mu}})$ and calculating a subgradient for it involve maximizing $F(\boldsymbol{\mu}; \boldsymbol{\theta})$ over $\boldsymbol{\mu}$.[4] Since $F(\boldsymbol{\mu}; \boldsymbol{\theta})$ is not concave in the general case, this appears to be a hard problem.[5]

The common practice in this case is to find a stationary point of $F(\boldsymbol{\mu}; \boldsymbol{\theta})$ using BP [26] and using it as a subgradient in subgradient ascent. Although it is hard to obtain theoretical guarantees for such an approach, it sometimes provides good empirical results [14, 16].

Another common approach comes from studying the tree graph case. In that case, the maximum likelihood parameters for $\bar{\boldsymbol{\mu}}$ are given in closed form by $\boldsymbol{\theta}^c(\bar{\boldsymbol{\mu}})$ defined as [21]:

$$\theta_i^c(x_i; \bar{\boldsymbol{\mu}}) = \log \bar{\mu}_i(x_i)$$
$$\theta_{ij}^c(x_i, x_j; \bar{\boldsymbol{\mu}}) = \log \frac{\bar{\mu}_{ij}(x_i, x_j)}{\bar{\mu}_i(x_i) \bar{\mu}_j(x_j)}$$

When the graph is not a tree, $\boldsymbol{\theta}^c(\bar{\boldsymbol{\mu}})$ is not necessarily the maximum likelihood parameter, as we also show in what follows. However, it is always true that $\bar{\boldsymbol{\mu}}$ is a stationary point of $F(\boldsymbol{\mu}; \boldsymbol{\theta}^c(\bar{\boldsymbol{\mu}}))$ [21]. Specifically, if one initializes BP on the MRF $\boldsymbol{\theta}^c(\bar{\boldsymbol{\mu}})$ with uniform messages, $\bar{\boldsymbol{\mu}}$ will be obtained as a fixed point (although this fixed point may be unstable as we show later).

---

[4]This maximization is subject to constraint $\boldsymbol{\mu} \in \mathcal{M}_L$. In what follows we do not state this explicitly for brevity.

[5]We are not aware of complexity results that prove this fact.

## 2 Bethe Learnable Parameters

Our main goal in the current work is to understand when Bethe learning achieves moment matching. By moment matching we mean that inference on the learned parameter will result in the empirical marginals (as in Eq. 3). In the context of Bethe learning, it is natural to define inference as returning the maximizer of $F(\boldsymbol{\mu}; \boldsymbol{\theta})$.[6] One difficulty with this is that $F$ is hard to maximize due to its non-convexity.[7] However, there is a more fundamental difficulty: $F(\boldsymbol{\mu}; \boldsymbol{\theta}_B(\bar{\boldsymbol{\mu}}))$ might have multiple distinct global maximizers. In this case, the outcome of the inference step is not well defined, and hence moment matching cannot be achieved. Furthermore, in this case the empirical marginals $\bar{\boldsymbol{\mu}}$ will generally not correspond to *any* of the maximizers of $F(\boldsymbol{\mu}; \boldsymbol{\theta}_B(\bar{\boldsymbol{\mu}}))$, but will rather lie in their convex hull (see Section 1.2). On the other hand, if $F(\boldsymbol{\mu}; \boldsymbol{\theta}_B(\bar{\boldsymbol{\mu}}))$ has a single global maximum then by Eq. 9 this maximizer is exactly $\bar{\boldsymbol{\mu}}$ and we have moment matching.

This brings us to our key question: for which values of $\bar{\boldsymbol{\mu}}$ will Bethe achieve moment matching? Given the above discussion, these are $\bar{\boldsymbol{\mu}}$ for which the optimal $\boldsymbol{\theta}_B(\bar{\boldsymbol{\mu}})$ has a single maximum.

**Definition 1.** *$\bar{\boldsymbol{\mu}}$ is Bethe learnable if there exists a $\boldsymbol{\theta}$ such that $\bar{\boldsymbol{\mu}}$ is the unique maximizer of $F(\boldsymbol{\mu}; \boldsymbol{\theta})$. In this case $\boldsymbol{\theta}$ maximizes the Bethe likelihood (i.e., $\boldsymbol{\theta} = \boldsymbol{\theta}_B(\bar{\boldsymbol{\mu}})$), and $\bar{\boldsymbol{\mu}}$ can be recovered from $\boldsymbol{\theta}$ so that moment matching is achieved. Denote the set of such $\bar{\boldsymbol{\mu}}$ by $\mathcal{B}_\mathcal{L}$.*

The definition of $\mathcal{B}_\mathcal{L}$ is quite implicit and it is thus not immediately clear what can be said about this set. For example, are there parameters that are not Bethe learnable? Clearly there are parameters that *are* Bethe learnable since Bethe is exact for trees (see Section 1.3). In what follows we show that there are in fact $\bar{\boldsymbol{\mu}}$ that are not in $\mathcal{B}_\mathcal{L}$ and characterize those in some cases.

## 3 Characterizing Unlearnable Marginals

The naive approach to testing if $\bar{\boldsymbol{\mu}} \in \mathcal{B}_\mathcal{L}$ would be to scan all parameters values $\boldsymbol{\theta}$, and for each of those test if $\bar{\boldsymbol{\mu}}$ is the unique maximizer of $F(\boldsymbol{\mu}; \boldsymbol{\theta})$. In what follows we provide several simpler approaches, some in closed form. Specifically, Sections 3.1 and 3.2 provide outer bounds on $\mathcal{B}_\mathcal{L}$ and Section 3.3 provides an inner bound.

### 3.1 A Condition Using Canonical Parameters

Our first observation in terms of characterizing $\mathcal{B}_\mathcal{L}$ is a lemma that provides a sufficient condition for $\bar{\boldsymbol{\mu}} \notin \mathcal{B}_\mathcal{L}$. In other words, we obtain an outer bound on $\mathcal{B}_\mathcal{L}$.

**Lemma 1.** *Let $\bar{\boldsymbol{\mu}} \in \mathcal{M}_L$ and let $\boldsymbol{\theta}^c(\bar{\boldsymbol{\mu}})$ be its canonical parameter. If $\bar{\boldsymbol{\mu}}$ is not a global maximizer of $F(\boldsymbol{\mu}; \boldsymbol{\theta}^c(\bar{\boldsymbol{\mu}}))$ then $\bar{\boldsymbol{\mu}} \notin \mathcal{B}_\mathcal{L}$.*

*Proof.* We are given the fact that:

$$\bar{\boldsymbol{\mu}} \notin \arg\max_{\boldsymbol{\mu} \in \mathcal{M}_L} F(\boldsymbol{\mu}; \boldsymbol{\theta}^c(\bar{\boldsymbol{\mu}})) \qquad (10)$$

Asssume by contradiction that $\bar{\boldsymbol{\mu}} \in \mathcal{B}_\mathcal{L}$ so that there is a parameter $\tilde{\boldsymbol{\theta}}$ for which $\bar{\boldsymbol{\mu}} \in \arg\max_{\boldsymbol{\mu} \in \mathcal{M}_L} F(\boldsymbol{\mu}; \tilde{\boldsymbol{\theta}})$. As mentioned in Section 1.3 $\bar{\boldsymbol{\mu}}$ is a fixed point of BP for the parameters $\boldsymbol{\theta}^c(\bar{\boldsymbol{\mu}})$. Since $\bar{\boldsymbol{\mu}}$ maximizes the negative Bethe free energy for $\tilde{\boldsymbol{\theta}}$, it is also a BP fixed point for $\tilde{\boldsymbol{\theta}}$ [26, Theorem 3]. Since BP fixed points correspond to re-parameterizations of the MRF [20], and since both $\boldsymbol{\theta}^c(\bar{\boldsymbol{\mu}})$ and $\tilde{\boldsymbol{\theta}}$ have the same BP fixed point $\bar{\boldsymbol{\mu}}$, it follows that $\boldsymbol{\theta}^c(\bar{\boldsymbol{\mu}})$ and $\tilde{\boldsymbol{\theta}}$ are re-parametrizations of the same MRF. It is then easy to show that two parameters that are re-parametrizations of each other share the same BP fixed points with the same values up to a constant. This implies that $\bar{\boldsymbol{\mu}}$ does in fact maximize $F(\boldsymbol{\mu}; \boldsymbol{\theta}^c(\bar{\boldsymbol{\mu}}))$, contradicting our assumption in Eq. 10. □

Lemma 1 provides an easy procedure for excluding points from $\mathcal{B}_\mathcal{L}$: Given $\bar{\boldsymbol{\mu}}$ generate the canonical parameter $\boldsymbol{\theta}^c(\bar{\boldsymbol{\mu}})$. Run BP with this parameter multiple times (e.g., by initializing messages randomly). If a fixed point with value $F(\boldsymbol{\mu}; \boldsymbol{\theta}^c(\bar{\boldsymbol{\mu}}))) > F(\bar{\boldsymbol{\mu}}; \boldsymbol{\theta}^c(\bar{\boldsymbol{\mu}}))$ is found then $\bar{\boldsymbol{\mu}} \notin \mathcal{B}_\mathcal{L}$.

Another interesting implication of Lemma 1 (by negation) is that if $\bar{\boldsymbol{\mu}}$ is learnable then its maximum likelihood parameter is necessarily the canonical one $\boldsymbol{\theta}^c(\bar{\boldsymbol{\mu}})$. Thus, the canonical parameters are in some sense the best choice for maximum likelihood with a Bethe approximation. Namely, they are the correct maximum likelihood parameters when $\bar{\boldsymbol{\mu}}$ is learnable. When $\bar{\boldsymbol{\mu}}$ is not learnable using Bethe approximation in learning is apparently not a good approach in any case.

### 3.2 Closed Form Bounds via the Hessian

Here we provide another outer bound on $\mathcal{B}_\mathcal{L}$ (i.e., a criterion for $\boldsymbol{\mu} \notin \mathcal{B}_\mathcal{L}$). The key idea is to find marginals $\bar{\boldsymbol{\mu}}$ which can *never* be a maximum (local or global) of $F(\boldsymbol{\mu}; \boldsymbol{\theta})$ *regardless of the value of $\boldsymbol{\theta}$*. Clearly, such $\bar{\boldsymbol{\mu}}$ will not be in $\mathcal{B}_\mathcal{L}$.

---

[6] Using the same inference approximation in learning and inference is also motivated by the results in [19, 21] where a similar approach for convex free energies results in moment matching.

[7] In practice, maximization will be approximated by running BP on the MRF given by $\boldsymbol{\theta}$.

For this purpose we will focus on binary pairwise models. Following [24, 10] we will use a minimal parameterization. In this representation we denote $\mu_i(x_i = 1)$ as $\mu_i$ and $\mu_{ij}(x_i = 1, x_j = 1)$ as $\mu_{ij}$. The other singleton and pairwise marginals can be obtained from these via:

$$\begin{aligned}\mu_{ij}(x_i = 1, x_j = 0) &= \mu_i - \mu_{ij} \\ \mu_{ij}(x_i = 0, x_j = 1) &= \mu_j - \mu_{ij} \\ \mu_{ij}(x_i = 0, x_j = 0) &= \mu_{ij} - \mu_i - \mu_j + 1 \\ \mu_i(x_i = 0) &= 1 - \mu_i\end{aligned}$$

It can be verified that a point is in $\mathcal{M}_L(G)$ if and only if all the above marginals are non-negative (e.g., see [3, 17]). The function $F(\boldsymbol{\mu}; \boldsymbol{\theta})$ in this case can be expressed as a function of $\mu_{ij}, \mu_i$ alone. Furthermore, there is no need to add non-negativity constraints since these would result in log of a negative number in the objective. For this unconstrained $F(\boldsymbol{\mu}; \boldsymbol{\theta})$, a sufficient condition for $\bar{\boldsymbol{\mu}}$ *not* to be a maximum for any $\boldsymbol{\theta}$ is that the Hessian of $F(\boldsymbol{\mu}; \boldsymbol{\theta})$ at $\bar{\boldsymbol{\mu}}$ is *not* negative-semi-definite regardless of the value of $\boldsymbol{\theta}$. Luckily, the Hessian at $F(\boldsymbol{\mu}; \boldsymbol{\theta})$ does not depend on $\boldsymbol{\theta}$ (since $F$ is linear in $\boldsymbol{\theta}$) and equals the Hessian of the Bethe entropy $H_B(\{\mu_{ij}, \mu_i\})$. Thus we have:

**Lemma 2.** *For binary variables, if the Hessian of $H_B(\{\mu_{ij}, \mu_i\})$ at $\boldsymbol{\mu} = \bar{\boldsymbol{\mu}}$ is not negative-semi-definite then $\bar{\boldsymbol{\mu}} \notin \mathcal{B}_\mathcal{L}$.*

Lemma 2 provides a sufficient condition for $\bar{\boldsymbol{\mu}} \notin \mathcal{B}_\mathcal{L}$. It can be easily tested for any given $\bar{\boldsymbol{\mu}}$ by calculating the Hessian at this point and checking its eigenvalues. To better understand the condition on the Hessian, we turn to a more specific scenario which results in a closed form expression on $\bar{\boldsymbol{\mu}}$.

We focus on marginals $\bar{\boldsymbol{\mu}}$ that are homogenous in the sense that all pairwise $\mu_{ij} = \mu_e$ for a constant $\mu_e$ and all $\mu_i = \mu_v$ for a constant $\mu_v$.[8] We make no restrictions on the graph structure. The following lemma states a simple sufficient condition for guaranteeing that $\bar{\boldsymbol{\mu}} \notin \mathcal{B}_\mathcal{L}$. We denote by $N_V$ the number of variables and $N_E$ the number of edges.

**Lemma 3.** *Assume $\bar{\boldsymbol{\mu}}$ corresponds to homogenous binary marginals in minimal representation, and $\bar{\boldsymbol{\mu}}$ satisfies the following condition:*

$$\bar{\mu}_e > \frac{(1 - \frac{N_V}{N_E})\bar{\mu}_v^2 + \frac{N_V}{2N_E}\bar{\mu}_v}{1 - \frac{N_V}{2N_E}} \quad (11)$$

*Then the Hessian of $H_B(\boldsymbol{\mu})$ at $\bar{\boldsymbol{\mu}}$ is not negative-semi-definite, and hence $\bar{\boldsymbol{\mu}} \notin \mathcal{B}_\mathcal{L}$.*

---
[8]Note that $\mathcal{M}_L(G)$ in this case reduces to the constraints: $2\mu_v - 1 \leq \mu_e \leq \mu_v$ and $\mu_e \geq 0$.

The proof of the lemma is given in Appendix A. This result has several interesting implications:

- The condition is independent of the graph structure. It only depends on the number of variables and edges.

- For tree graphs and single loop graphs there are no $\bar{\boldsymbol{\mu}}$ that satisfy the conditions. This is consistent with the fact that the Bethe free energy has a unique optimum in these cases and all marginals are Bethe learnable.

- As $\frac{N_V}{N_E} \to 0$ (e.g., for complete graphs with $N_V \to \infty$) the condition becomes $\bar{\mu}_e \geq \bar{\mu}_v^2$. Perhaps surprisingly, this is a condition that is satisfied by any ferromagnetic distribution [15, 4].[9] This implies that if $\bar{\boldsymbol{\mu}}$ are generated from a ferromagnetic distribution with $\frac{N_V}{N_E} \to 0$, then $\bar{\boldsymbol{\mu}} \notin \mathcal{B}_\mathcal{L}$. In other words, ferromagnets are not Bethe learnable in this asymptotic regime.

### 3.3 Inner Bounds on $\mathcal{B}_\mathcal{L}$

The results in the previous sections provide outer bounds on $\mathcal{B}_\mathcal{L}$. In the current section we show how previous results on BP convergence can be utilized to obtain *inner* bounds on $\mathcal{B}_\mathcal{L}$. The key idea is to check, given some $\bar{\boldsymbol{\mu}}$, whether the canonical parameters $\boldsymbol{\theta}^c(\bar{\boldsymbol{\mu}})$ yield moment matching. We know that $\bar{\boldsymbol{\mu}}$ is a BP fixed point for $\boldsymbol{\theta}^c(\bar{\boldsymbol{\mu}})$. Thus a sufficient condition for moment matching to take place is that $\bar{\boldsymbol{\mu}}$ is the *unique* stable fixed point of BP.[10] This would imply that at $\boldsymbol{\theta}^c(\boldsymbol{\mu})$ we have moment matching for $\bar{\boldsymbol{\mu}}$ and thus $\bar{\boldsymbol{\mu}} \in \mathcal{B}_\mathcal{L}$.

Fortunately there are various results that provide sufficient conditions on a parameter $\boldsymbol{\theta}$ having a unique stable fixed point. This has been addressed by multiple works in the past (e.g., [11, 13]).[11] Each of these works provides a different condition on $\boldsymbol{\theta}$. We are not concerned with which one is better since all can equally well be applied to our case and we can simply take the union of the conditions to obtain a tighter bound.

Thus, our condition for $\bar{\boldsymbol{\mu}} \in \mathcal{B}_\mathcal{L}$ is calculated as follows: for a given $\bar{\boldsymbol{\mu}}$, calculate $\boldsymbol{\theta}^c(\bar{\boldsymbol{\mu}})$. Now check whether $\boldsymbol{\theta}^c(\bar{\boldsymbol{\mu}})$ has a unique BP fixed point (by using one of the conditions in the papers above). If it does, then $\bar{\boldsymbol{\mu}} \in \mathcal{B}_\mathcal{L}$. In the experiments section we will specifically look at the condition from [13].

---
[9]The non-homogenous form is $\mu_{ij} \geq \mu_i\mu_j$.
[10]Stability is needed to guarantee that this point is in fact a global maximum as our analysis requires [6].
[11]Note also [7] which analyzes unique fixed points, but not necessarily stable.

## 4 Related Work

Several works have analyzed the behavior of Bethe approximations in graphical models, and their relation to BP. The first key work is [26], which showed that the fixed points of BP are local optima of the Bethe free energy. Later this result was refined in [6] to show that the stable fixed points of BP are local *minima* of the Bethe free energy.

Another line of work focuses on conditions under which the BP has a unique fixed point (implying no local minima via [6]). Such works (e.g., [18, 11, 13, 7]) typically provide sufficient conditions on the model structure and parameters for BP to have a unique fixed point. As we show in Section 3.3, such results can be used to obtain inner bounds on $\mathcal{B}_\mathcal{L}$, although they were not originally developed for this purpose.

There has been less work on understanding learning with Bethe approximations. The canonical parameters (see Section 1.3) have been suggested in several works [21, 23, 25]. As we show here, these parameters are generally not the ones optimizing the Bethe likelihood, but when $\bar{\boldsymbol{\mu}}$ is learnable, they are in fact optimal. Much stronger theory is available for learning with convex free energies such as tree-reweighted variants. In [19] it is shown that such methods have desirable stability and asymptotic properties. The performance of Bethe learning is not analyzed in this context, since it does not fall under the convex approximations.

Our work proposes a novel view of learning with Bethe, which is to focus on the marginals that Bethe can and cannot match during learning. This highlights the regimes where Bethe cannot be expected to work well, and those where it might work, as we show further in our empirical results. Our results provide initial characterization of the set $\mathcal{B}_\mathcal{L}$ in terms of inner and outer bounds. We expect that these can be tightened further.

## 5 Illustrating the Bounds

In this section we provide several graphical illustrations of the bounds on $\mathcal{B}_\mathcal{L}$ that we presented earlier. We focus on the case of a two dimensional $3 \times 3$ toroidal grid graph, and on homogenous parameters as described in Section 3.2. In this case, $\bar{\boldsymbol{\mu}}$ can be conveniently represented in two dimensions (i.e., $\mu_v, \mu_e$).

We begin by showing a case where the empirical marginals $\bar{\boldsymbol{\mu}}$ are unlearnable. In this case, the maximum Bethe likelihood parameter $\theta_B(\bar{\boldsymbol{\mu}})$ results in a function $F(\boldsymbol{\mu}; \theta_B(\bar{\boldsymbol{\mu}}))$ which is *not* maximized by $\bar{\boldsymbol{\mu}}$ (i.e., moment matching is not achieved). Rather, the function $F$ has two other maxima, and $\bar{\boldsymbol{\mu}}$ lies in their convex hull, as described in Section 1.2. This is shown in Figure 1.

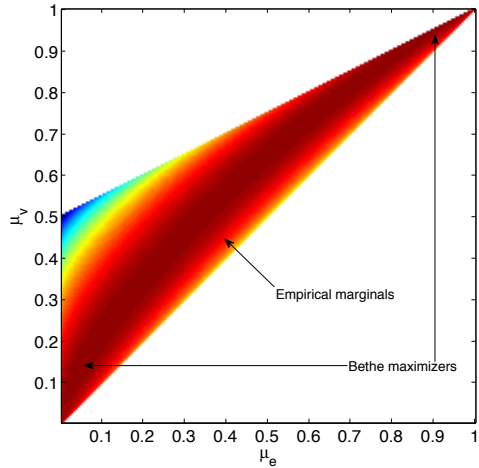

Figure 1: Illustration of a case where $\bar{\boldsymbol{\mu}}$ (denoted by empirical marginals in the figure) does not maximize $F(\boldsymbol{\mu}; \theta_B(\bar{\boldsymbol{\mu}}))$. However, $\bar{\boldsymbol{\mu}}$ is obtained as a convex combination of the two global maxima of $F(\boldsymbol{\mu}; \theta_B(\bar{\boldsymbol{\mu}}))$. The colormap corresponds to the values of $F(\boldsymbol{\mu}; \theta_B(\bar{\boldsymbol{\mu}}))$. In this case $\theta_B(\bar{\boldsymbol{\mu}})$ was found by exhaustive search.

Figure 2 depicts several results regarding the set $\mathcal{B}_\mathcal{L}$. The colored regions correspond to marginals which can be obtained via *some* empirical distribution (i.e., the marginal polytope [23]). The blue region indicates marginals that are learnable according to the inner bound in Section 3.3. The red region indicates marginals that are unlearnable according to the outer bounds in Sections 3.1 and 3.2. In this case, lemmas 1, 2 and 3 yield identical outer bounds. The region in black is not covered by any of our bounds. However, there is an easy way to check empirically whether it might be learnable: run gradient descent on the Bethe likelihood (using BP to approximate the objective and gradient[12]), and check whether the parameters at convergence satisfy moment matching. It turns out that for the region in black, moment matching is achieved. Note that due to the potential suboptimality of BP, this is not a theoretical guarantee that this region is indeed learnable (i.e., it could be that we have reached a suboptimal parameter, or that the reported marginals are not the optimal ones for Bethe). Taken together, the results in Figure 2 suggest that our outer bounds are tight for the given graph.

## 6 Discussion

This work presents an analysis of when learning with Bethe is guaranteed to fail, in the sense of not match-ing the moments of the training data. One of the key difficulties of the Bethe approximation of marginals is the non-convexity of the Bethe free energy, resulting in local optima during inference. Our results show that when using Bethe within learning, another problem surfaces, even in the case where exact optimization of the Bethe free energy is possible (and hence learning is tractable since the Bethe likelihood is concave). Namely, that for particular empirical marginals (those outside $\mathcal{B}_\mathcal{L}$) moment matching will not be achieved.

---

[12]This is an approximation since BP does not necessarily find the global optimum of the Bethe free energy.

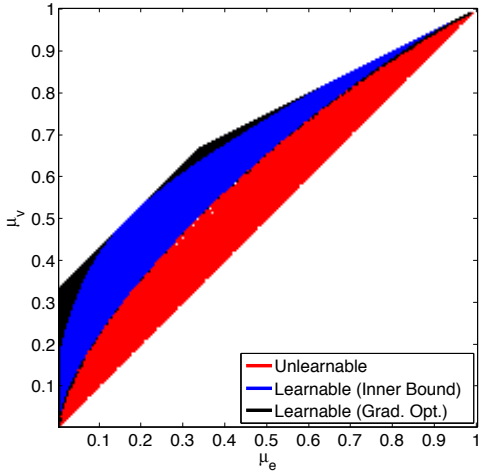

Figure 2: Illustration of bounds on $\mathcal{B}_\mathcal{L}$ space. Each point in the figure corresponds to an empirical marginal $\bar{\boldsymbol{\mu}}$. The blue region shows marginals that are learnable according to Section 3.3. The red region shows marginals that are unlearnable according to Sections 1.3 and 3.2. The black region is not covered by any of our bounds. However, when running gradient ascent on the Bethe likelihood with these marginals, it turns out that moment matching is achieved at convergence (we decide that moment matching is achieved when the absolute difference between empirical and Bethe marginals is less than 0.01).

Characterizing the set $\mathcal{B}_\mathcal{L}$ is a challenge, and we have presented initial steps in this direction by showing how to obtain inner and outer bounds on it, some in closed form. Our analysis also highlights the fact that the canonical parameters often used in learning are inadequate in some cases. On the other hand, they can be used to obtain outer bounds on $\mathcal{B}_\mathcal{L}$ as in Lemma 1.

Many interesting questions arise from our analysis and deserve further study: e.g., when are the inner and outer bounds we presented tight, and whether tighter bounds exist? When are the canonical parameters optimal and in $\mathcal{B}_\mathcal{L}$?

On a practical level our results imply that there are regimes when Bethe learning is bound to fail, and that in some cases they can be inferred from the empirical marginals without running a learning algorithm (e.g., by checking one of our outer bounds). Additionally, we show that in the binary homogenous models, there are graphs where ferromagnetic models are unlearnable.

Our analysis of the Hessian in Section 3.2 shows that there are marginals $\bar{\boldsymbol{\mu}}$ that cannot be local minima of the Bethe free energy. This in turn implies that they will not be stable fixed points of BP. This highlights a very interesting limitation of using BP to approximate marginals. Namely, that there regions of marginals space which will never be obtained as a result of running BP. It will be interesting to study the practical implications of this observations.

Finally, our results do not imply that one should not use Bethe approximations within learning. It has been previously shown that Bethe approximations of marginals outperform convex variants across a wide range of parameter settings [9]. It is thus certainly possible that for data such that $\bar{\boldsymbol{\mu}} \in \mathcal{B}_\mathcal{L}$ the learned parameters will perform well. We intend to address this issue as well as other theoretical questions in future work.

## A  Proof of Lemma 3

We first note that the Hessian of $H_B(\boldsymbol{\mu})$ has a particularly simple form in the homogeneous case. Denote this Hessian matrix by $A$. Then its elements correspond to only $N_V+3$ unique numbers $a_i$ ($i=1,\ldots,V$), $b$, $c$, $d$, which are given by:

$$\begin{aligned}
a_i &= (d_i - 1)(\frac{1}{\mu_v} + \frac{1}{1-\mu_v}) - d_i c \\
b &= -\frac{1}{1-2\mu_v+\mu_e} \\
c &= \frac{1}{(\mu_v-\mu_e)} + \frac{1}{(1-2\mu_v+\mu_e)} \\
d &= -\frac{1}{\mu_e} - \frac{1}{\mu_v-\mu_e} - c
\end{aligned}$$

The Hessian depends on these values via:

$$A_{[k,l]} = \begin{cases} a_k & k \in V, l \in V & l = k \\ b & k \in V, l \in V & l \in N(k) \\ c & k \in V, l \in E & k \in l \\ c & k \in E, l \in V & l \in k \\ d & k, l \in E & k = l \\ 0 & otherwise \end{cases} \quad (12)$$

where $N(k)$ is the set of neighbors of node $k$. The indexing scheme is as follows: by $k \in V$ we mean $k$ is the coordinate corresponding to $\mu_k$ and $k \in E$ means $k$ is the coordinate corresponding to some edge in $E$.

The Hessian A is negative-semi-definite iff for all $\boldsymbol{z}$ it holds that $\boldsymbol{z}^T A \boldsymbol{z} \leq 0$. We will show that if the condition in Eq. 11 is satisfied then we can find a $\boldsymbol{z}$ such that $\boldsymbol{z}^T A \boldsymbol{z} > 0$ and therefore $A$ is not negative definite, and the lemma follows. We will define such a $\boldsymbol{z}$ in the following way: $z_i = 1$ if $i \in V$ and $z_i = z$ (for some scalar $z$) if $i \in E$. Denoting $\hat{a} = \frac{1}{\mu_v} + \frac{1}{1-\mu_v}$ we obtain after some algebra that:[13]

$$\boldsymbol{z}^T A \boldsymbol{z} = (2E - V)\hat{a} - 2N_E c + 2N_E b + N_E z^2 d + 4N_E z c$$

The above is a quadratic concave function in $z$. When its discriminant is greater than zero, it will attain positive values, and A will not be negative definite. The condition on the discriminant corresponds to:

$$0 < c^2 - \frac{1}{2}d\left(\hat{a} - c + b\right) + \frac{N_V}{4N_E}d\hat{a} \qquad (13)$$

Assigning the values of the Hessian and some more algebra leads to the equation:

$$0 < (\mu_e - \mu_v^2)(1 - \frac{N_V}{2N_E}) - \frac{N_V}{2N_E}\mu_v(1-\mu_v) \qquad (14)$$

Switching sides we get the condition in Eq. 11. $\square$

---

[13] Because of the structure of the matrix elements, the dependence on the degrees $d_i$ nicely cancels out after summation.